\documentclass[5p]{elsarticle}

\usepackage{dblfloatfix}

\usepackage{times}
\usepackage[colorlinks=true]{hyperref}

\usepackage{graphicx}

\usepackage{soul}
\usepackage{color}
\sethlcolor{yellow}

\usepackage{booktabs}
\usepackage{multirow}

\journal{Fusion Engineering and Design}

\begin{document}

\begin{frontmatter}

\title{Deep learning for plasma tomography using the bolometer system at JET}

\author[ist]{Francisco A. Matos}

\author[ist]{Diogo R. Ferreira\corref{cor}}
\ead{diogo.ferreira@tecnico.ulisboa.pt}
\cortext[cor]{Corresponding author}

\author[ipfn]{Pedro J. Carvalho}

\author[]{JET Contributors\fnref{fn}}
\fntext[fn]{See the Appendix of F. Romanelli et al., Proceedings of the 25th IAEA Fusion Energy Conference 2014, Saint Petersburg, Russia}

\renewcommand{\elsaddress}{EUROfusion Consortium, JET, Culham Science Centre, Abingdon, OX14 3DB, UK\\}
\address[ist]{Instituto Superior T\'{e}cnico (IST), University of Lisbon, Portugal}
\address[ipfn]{Instituto de Plasmas e Fus\~{a}o Nuclear (IPFN), IST, University of Lisbon, Portugal}

\begin{abstract}
Deep learning is having a profound impact in many fields, especially those that involve some form of image processing. Deep neural networks excel in turning an input image into a set of high-level features. On the other hand, tomography deals with the inverse problem of recreating an image from a number of projections. In plasma diagnostics, tomography aims at reconstructing the cross-section of the plasma from radiation measurements. This reconstruction can be computed with neural networks. However, previous attempts have focused on learning a parametric model of the plasma profile. In this work, we use a deep neural network to produce a full, pixel-by-pixel reconstruction of the plasma profile. For this purpose, we use the overview bolometer system at JET, and we introduce an up-convolutional network that has been trained and tested on a large set of sample tomograms. We show that this network is able to reproduce existing reconstructions with a high level of accuracy, as measured by several metrics.
\end{abstract}


\end{frontmatter}


\section{Introduction}

Tomography \cite{anton96xray,ingesson98tomography} is a plasma diagnostics technique which can be used to determine the plasma profile on a poloidal cross-section of the fusion device. This 2D plasma profile is obtained through a reconstruction process based on the measurements of radiation detectors placed around the plasma, such as X-ray cameras and bolometer systems \cite{ingesson08tomography}. The reconstruction process is often referred in the literature as tomographic inversion \cite{mlynar10inversion}.

There are several methods for tomographic inversion, including the maximum entropy method, the Tikonov regularization method, and the minimum Fisher information method, among others \cite{kim06comparison,craciunescu09comparison}. On the other hand, there have been several attempts to produce tomographic reconstructions with neural networks \cite{demeter97neural,barana02neural,ronchi10neural}. Once trained, a neural network can approximate the results of tomographic inversion with satisfactory accuracy, while being faster than the classical methods. This becomes especially useful when there are large amounts of tomographic data to be processed after a pulse or, more importantly, to enable the use of such reconstructions as input to the real-time control of the fusion experiment \cite{carvalho10control}.

All of these experiments with neural networks have been performed before the advent of deep learning \cite{nature15deeplearning,schmidhuber15deep}. In recent years, the field of neural networks has seen tremendous progress in several fronts, namely:
\begin{itemize}
\setlength\itemsep{0em}

\item the development of deep network architectures based on convolutional neural networks (CNNs) \cite{krizhevsky12imagenet,simonyan15verydeep} and recurrent neural networks (RNNs) \cite{graves12recurrent};

\item the use of new activation functions, especially the rectified linear unit (ReLU) \cite{nair10rectified} and its variants \cite{maas13rectifier,he15delving,clevert16elu};

\item new procedures for parameter initialization \cite{glorot10understanding,he15delving};

\item improvements in training algorithms \cite{duchi11adagrad,zeiler12adadelta,kingma15adam};

\item new methods to avoid overfitting, especially dropout \cite{srivastava14dropout} and batch normalization \cite{ioffe15batch};

\item the use of GPU computing (graphics processing units) to significantly accelerate network training \cite{raina09gpus,coates13deep};

\item and the development of new software libraries, such as Caffe \cite{jia14caffe}, Theano \cite{bergstra11theano} and Torch7 \cite{collobert11torch}, which made deep learning accessible to a wide range of applications.

\end{itemize}


These developments have led us to rethink the way neural networks can be used in plasma tomography. In particular, whereas previous attempts have focused on learning the parameters of a parametric model of the plasma profile \cite{ronchi10parametric}, here we aim at a full, pixel-by-pixel reconstruction of the 2D plasma profile with the same resolution as obtained by other tomography methods based on regularization. This is an interesting problem for deep learning, because it involves producing a 2D image from 1D projection data, while convolutional neural networks (CNNs) typically excel in doing the opposite, e.g.~turning a 2D image into a 1D vector of class probabilities that can be used for image classification.

In this work, we present an up-convolutional network to perform the reconstruction of the 2D plasma profile based on the overview bolometer system at JET, known as KB5 \cite{huber07upgraded}.
The network is trained on existing tomograms, and is able to reproduce previously unseen reconstructions with high accuracy, to the point that they are virtually indistinguishable from the original. To evaluate the quality of the results, we turn to several metrics that are commonly used for image comparison.

The paper is organized as follows: Section \ref{sec:kb5} describes the KB5 bolometer system and the data that has been collected from that diagnostic across a number of JET campaigns. Section \ref{sec:network} discusses data preprocessing, and presents the architecture of the up-convolutional network that has been devised for plasma tomography. Section \ref{sec:results} describes how the network was trained and tested, and discusses how the results have been evaluated using set of metrics. Finally, Section \ref{sec:conclusion} concludes the paper with a view towards the application of a similar approaches to other tomography diagnostics in JET and elsewhere.

\section{Bolometer systems and tomography at JET}
\label{sec:kb5}

The use of tomography diagnostics at JET dates back to the mid-1980s when the first bolometer system, named KB1, was installed \cite{mast85bolometric}. Originally, it included one vertical camera with 14 bolometers and two horizontal cameras, with 10 bolometers each, directed at the upper half and lower half of the vessel. 

This was followed by development of the KB3 and KB4 diagnostics in the mid-1990s. The KB3 bolometer system is mainly directed towards the divertor region at the bottom of the vessel. It has been the subject of several improvements over the years \cite{ingesson00characterization}, and it is still in use today. Currently, KB3 includes 7 cameras in the divertor region, each with 4 lines of sight. The KB4 diagnostic uses the same type of cameras, but it provides an overview of the entire plasma region. Originally, KB4 had a total of 14 cameras, but nowadays only 6 of them are operational, with only a few lines of sight. 

\begin{figure}[h]
\centering
\includegraphics[scale=0.09]{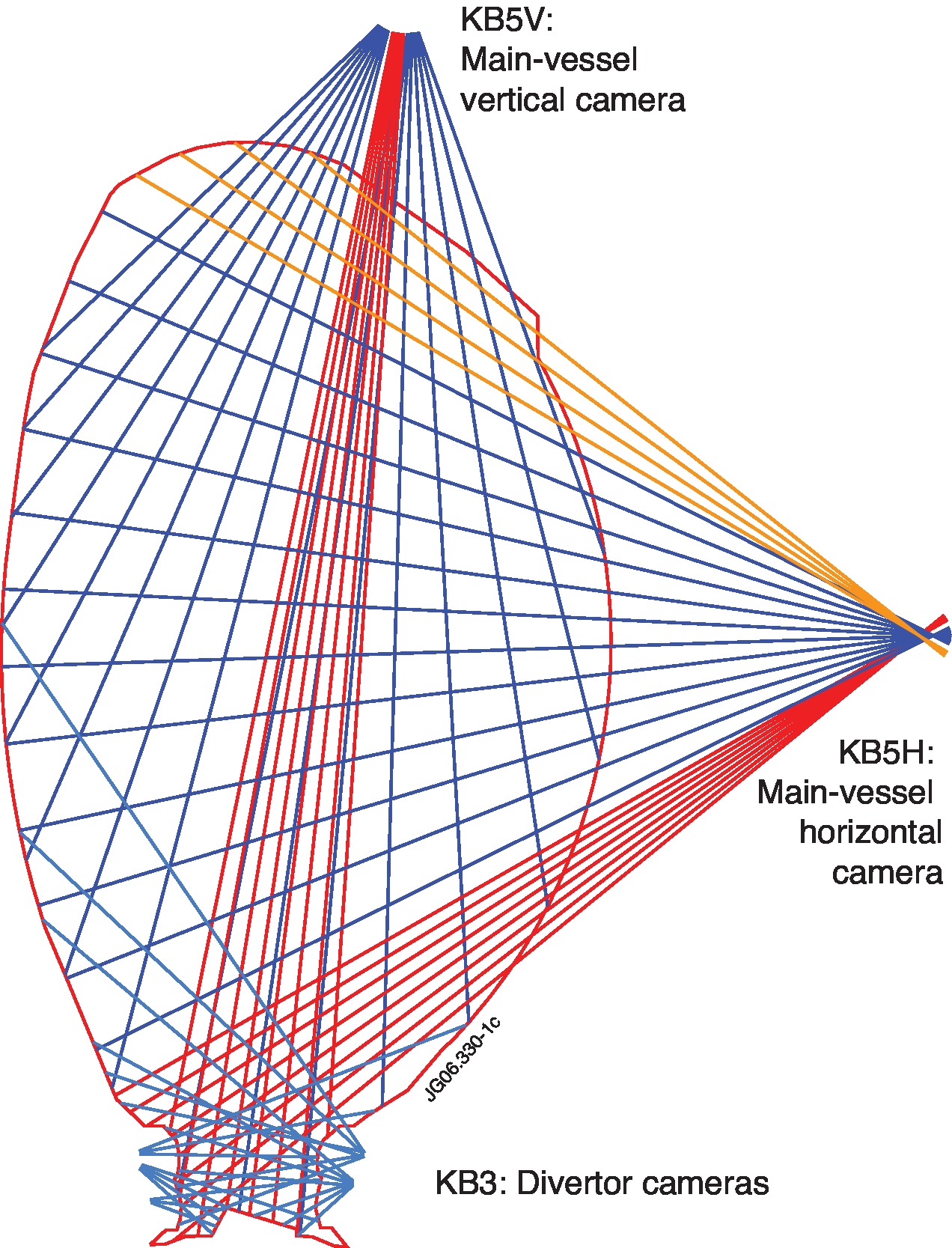}
\caption{KB5 cameras (EUROfusion figures database JG06.330-1c)}
\label{fig:kb5}
\end{figure}

In the late 1990s and early 2000s, a new bolometer system, named KB5, was developed \cite{mccormick05new,huber07upgraded,huber07improved}. The KB5 diagnostic comprises one horizontal camera (KB5H) and one vertical camera (KB5V) with 24 bolometers each. The lines of sight for each of these cameras are arranged in such a way that 16 channels cover the whole plasma, and 8 channels are directed towards the divertor region, as illustrated in Figure \ref{fig:kb5}. This means that KB5 provides an overview of the plasma region, and also a more detailed view of the divertor region.

\begin{figure}[h]
\centering
\includegraphics[scale=0.39]{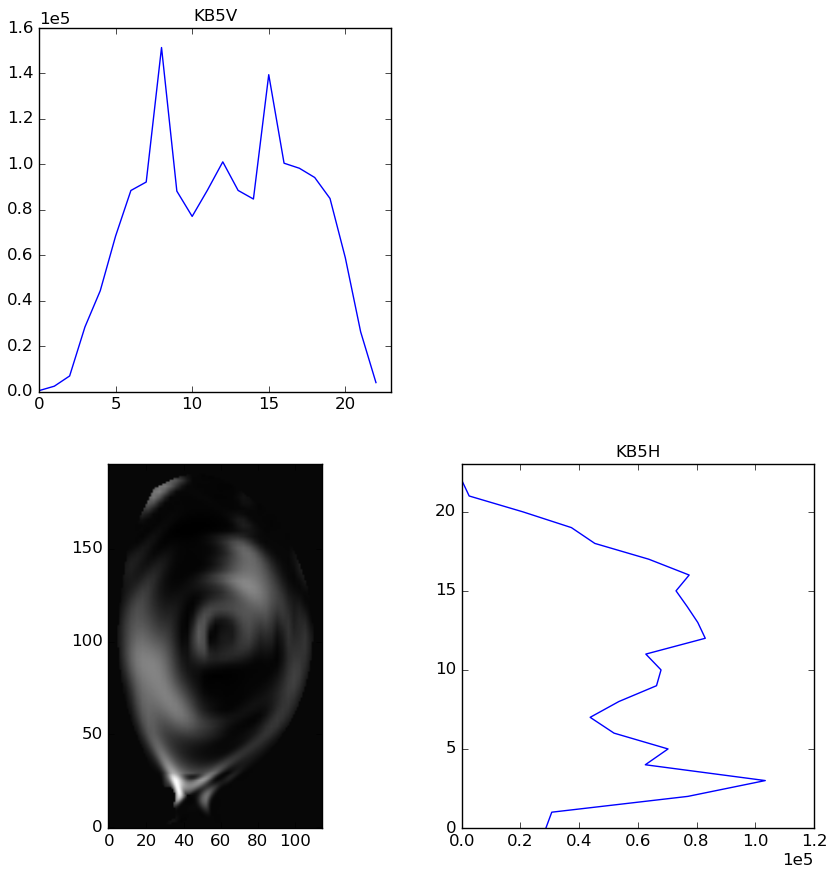}
\caption{KB5H/V readings and tomographic reconstruction for JET pulse no. 81850 at $t\!=\!51.35$ s.}
\label{fig:tomo}
\end{figure}

Figure \ref{fig:tomo} shows a sample reconstruction of the plasma profile from the readings of KB5H and KB5V. At JET, this kind of reconstruction is obtained through a tomographic method based on a series expansion of the plasma profile with a set of basis functions \cite{ingesson97radiation,ingesson00mathematics}. Thousands of such reconstructions have been computed over the years but, on average, only a few reconstructions are available per pulse. Therefore, to obtain a large number of reconstructions for training the network, it was necessary to gather data from many pulses.

\begin{figure*}[b]
	\centering
	\includegraphics[scale=0.55]{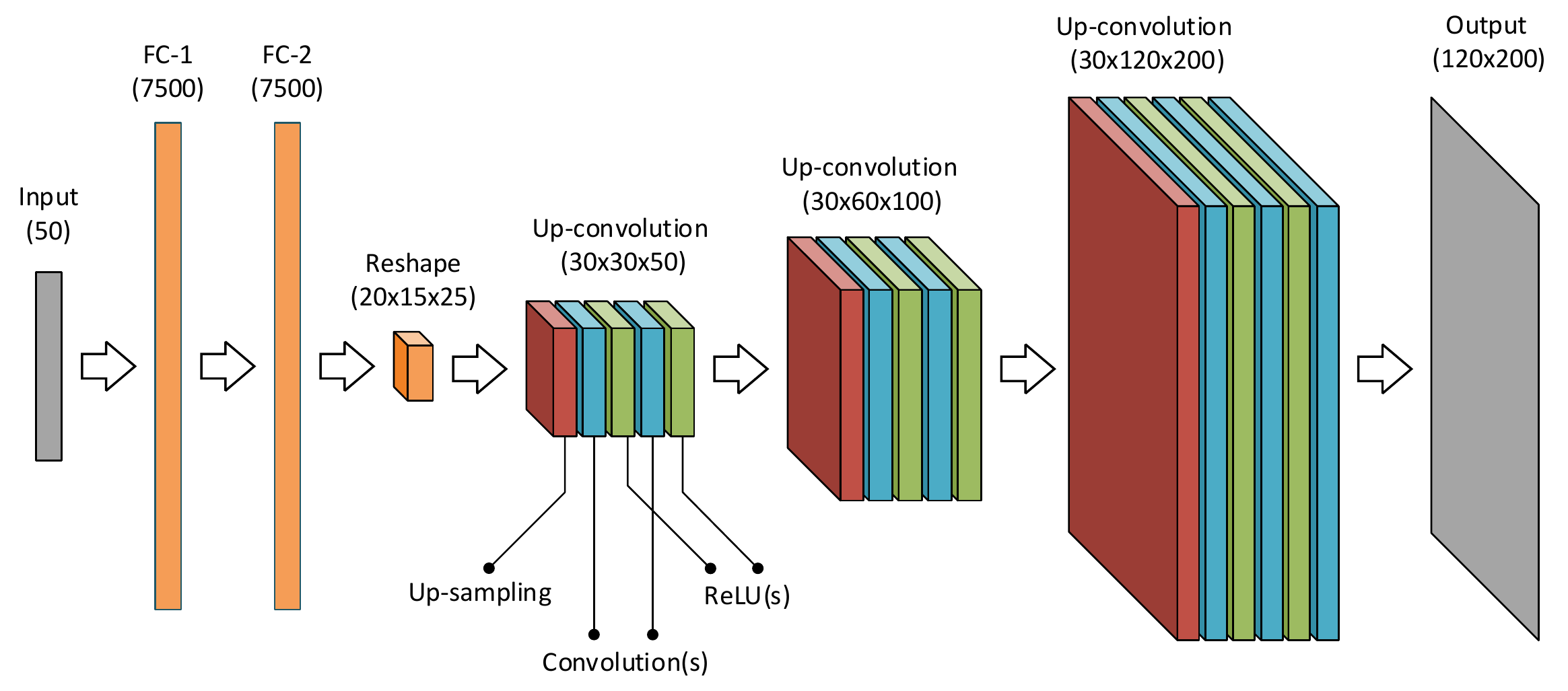}
	\caption{Architecture of the up-convolutional network}
	\label{fig:network}
\end{figure*}

In this work, we collected data for all JET campaigns since the installation of the ITER-like wall in September 2011 \cite{horton13wall}. Table \ref{tab:campaigns} shows the number of available reconstructions by campaign. In total, we gathered about 18\,000 sample reconstructions. However, there are significant differences between campaigns, with changes in operating conditions, diagnostics equipments, systems calibration, and even changes to the fuel itself (e.g.~C34 was a hydrogen campaign).

\begin{table}[h]
  \renewcommand{\arraystretch}{1.2} 
  \centering
  \scriptsize
  \begin{tabular}{lccccr}
    \hline
   	         & Start & Finish & First & Last  & \\
   	Campaign & date  & date   & pulse & pulse & Reconstructions\\
    \hline
	C28a & 2011-09-01 & 2011-09-20 & 80176 & 80372 &   41\\
	C28b & 2011-10-10 & 2011-10-28 & 80653 & 80976 &  375\\
	C28c & 2011-11-25 & 2011-12-21 & 81264 & 81643 &  282\\
	C29  & 2012-01-10 & 2012-04-10 & 81726 & 82905 & 3238\\
	C30b & 2012-05-15 & 2012-07-14 & 82944 & 83620 & 2255\\
	C30c & 2012-07-16 & 2012-07-27 & 83621 & 83794 &  590\\
	C31  & 2013-07-17 & 2013-09-27 & 84442 & 85355 & 3071\\
	C32  & 2013-09-30 & 2013-10-04 & 85356 & 85457 & 1091\\
	C32a & 2014-01-08 & 2014-01-10 & 85900 & 85978 &   11\\
	C33  & 2014-06-19 & 2014-09-05 & 86452 & 87583 & 4044\\
	C34  & 2014-09-10 & 2014-10-09 & 87584 & 87958 & 2134\\
	C35  & 2015-11-09 & 2015-12-18 & 88941 & 89472 &  633\\
	\hline
  \end{tabular}
  \caption{JET campaigns, dates, pulses and available reconstructions}
  \label{tab:campaigns}
\end{table}

Therefore, we divided the data into the following subsets:
\begin{itemize}
\setlength\itemsep{0em}
\item C29 with 3238 reconstructions;
\item C30b + C30c with 2845 reconstructions;
\item C31 + C32 with 4162 reconstructions;
\item C33 with 4044 reconstructions.
\end{itemize}

The idea was to merge campaigns only if they were very close in terms of dates, while at the same time trying to come up with datasets of roughly the same size. The network has been trained and tested separately on each of these datasets.

\section{The up-convolutional network}
\label{sec:network}

The KB5 system is able to collect readings at a sampling rate of 5 kHz. Each reading includes the 24 channels of KB5H, the 24 channels of KB5V, and up to 8 additional channels that are used as reserve \cite{mccormick05new}. In the data that we collected for this work, each reading had always 52 channels (24 + 24 + 4 reserve). Two of these channels were always zero, namely the last reserve channel and the first channel of KB5V. On the other hand, two channels of KB5V were known to be malfunctioning, and one of these was precisely the first channel. In any case, we did not discard any channels immediately. Instead, we applied Principal Component Analysis (PCA) \cite{jolliffe02pca} to determine how many channels were actually useful.

In deep learning, it is common practice to use some form of PCA to normalize the input data \cite{lee09unsupervised,ngiam11multimodal}. Usually, this is done mainly to reduce the dimensionality of the input. Here, because of the anomalies in some of the KB5 channels, we applied PCA to determine how many of them should be kept, and also to de-correlate the input data to make it easier for the network to learn. (It is interesting to note that PCA has been used in \cite{ronchi10neural} for this second purpose as well.)

By applying PCA decomposition to the input data coming from KB5, we found that 100\% of the variance could be explained with 50 components (rather than 52) so we transformed the input data and kept only those 50 components. The input to the network was therefore a 1D vector of length 50.

As for the output data, the available reconstructions had a resolution of 115$\times$196 pixels (as in Figure \ref{fig:tomo}). Our goal was to train the neural network on these pixel data so that it would produce images with the same resolution. However, to facilitate the design of the network architecture, we padded the images with zeros up to a resolution of 120$\times$200. This padding has no effect in the results, since the bordering pixels in the available reconstructions were already zero, and the network will learn that those extra outputs are always zero as well.

The challenge was therefore to devise a network architecture that receives a 1D input of length 50, and produces a 2D output of size 120$\times$200. This is in contrast with the architecture of most convolutional neural networks (CNNs), which typically receive a 2D image as input and produce a 1D vector of features or class probabilities as output. What we need here is the inverse of a CNN, a kind of de-convolutional network that receives a 1D vector and produces a 2D image.

The typical architecture of a CNN contains a series of convolutions, rectifiers, and pooling layers \cite{krizhevsky12imagenet,simonyan15verydeep}. Basically, the convolutions are intended to detect specific features, the rectifiers apply a threshold over those features, and the pooling layers perform image downsampling. These are then followed by a couple of fully-connected layers before the network output. Our idea was to turn this architecture around, by having the fully-connected layers at the beginning, and then a series of layers to perform de-convolution and unpooling to increase the image size up to the desired resolution.

The problem is that these inverse operations may be impossible to perform. For example, pooling (either max pooling or average pooling) loses information from its inputs, so there is no way to perform unpooling. In the literature, some authors were able to emulate unpooling by using additional information in the form of switch variables \cite{zeiler11adaptive,noh15deconvolution}. Here we do not have such information, so we followed a similar approach to \cite{dosovitskiy15learning}, which consists in performing upsampling (i.e.~doubling the size of the input image) followed by convolution. This is referred to as \emph{up-convolution}. The up-convolution is not the same as unpooling or de-convolution, but it can be used as an approximation to accomplish the same purpose.

\begin{table*}[b]
  \small
  \centering
  \begin{tabular}{@{}lcccccccccccc@{}}
    \toprule
      && \multicolumn{3}{c}{SSIM} && \multicolumn{3}{c}{PSNR} && \multicolumn{3}{c}{NRMSE} \\
    \cmidrule{3-5}
    \cmidrule{7-9}
    \cmidrule{11-13}
      && mean & best & worst && mean & best & worst && mean & best & worst \\
    \midrule

      C29 &&
	  0.978387 &   0.999324 &   0.901227 &&
	 37.102214 &  49.553505 &  27.362232 &&
	  0.033286 &   0.006915 &   0.089919 \\

      C30b + C30c &&
	  0.968798 &   0.986527 &   0.925779 &&
	 33.838852 &  36.898350 &  28.939540 &&
	  0.044555 &   0.030393 &   0.076322 \\

      C31 + C32 &&
	  0.990349 &   0.998765 &   0.953863 &&
	 36.222247 &  37.616395 &  29.413519 &&
	  0.032600 &   0.027523 &   0.070956 \\
      
      C33 &&
	  0.990571 &   0.999572 &   0.958346 &&
	 41.146799 &  47.095725 &  29.676929 &&
	  0.021336 &   0.009152 &   0.069047 \\

    \bottomrule
  \end{tabular}
    \caption{SSIM, PSNR, and NRMSE obtained on each test set}
  \label{tab:results}
\end{table*}

Figure \ref{fig:network} shows the architecture of the up-convolutional network that we devised for this work. The input layer is followed by two fully-connected (FC) layers with 7500 nodes, which are then reshaped into 20 feature maps\footnote{In a traditional CNN, a feature map is the result of a convolution.} of size 15$\times$25. The first up-convolutional layer enlarges these feature maps to 30$\times$50 and also increases their number to 30 through two convolutions and rectifiers. The second and third up-convolutional layers have a similar structure, and also work with 30 feature maps. These feature maps are successively enlarged until a final convolution merges them into a single image of 120$\times$200.

The size of the FC layers, the number of feature maps, and the number of convolutions have all been the subject of extensive experiments. However, we observed no significant improvements in the results by increasing them further. We also conducted a preliminary experiment where we trained the network with randomly-generated phantoms for which the projections could be calculated analytically. From the projections, the network was able to reproduce each phantom with an average error per pixel below 2\%.

\section{Training and results}
\label{sec:results}

To train and test the network, we partitioned each of the four datasets (C29, C30b + C30c, C31 + C32, and C33) into 80\% for training, 10\% for validation, and 10\% for testing. The validation set was used to monitor the training progress. Four separate training processes were carried out, one for each dataset, followed by testing on the corresponding test set.

The network was trained using stochastic gradient descent (SGD) \cite{lecun12backprop} with a relatively low learning rate of 0.001, which yielded better results in the long run than more sophisticated algorithms such as Adagrad \cite{duchi11adagrad} and Adadelta \cite{zeiler12adadelta}. The network weights were initialized with a Glorot-uniform distribution \cite{glorot10understanding}, and we used a batch size of 10 while training the network on the GPU (Nvidia Titan X). As loss function, we used the mean absolute error (MAE) between the original reconstructions and the images produced by the network. All of this was done using Python, Keras\footnote{Keras -- Deep Learning library for Theano and TensorFlow: \url{https://keras.io/}} and Theano\footnote{Theano library: \url{http://deeplearning.net/software/theano/}}.

\begin{figure}[h]
	\centering
	\includegraphics[scale=0.36]{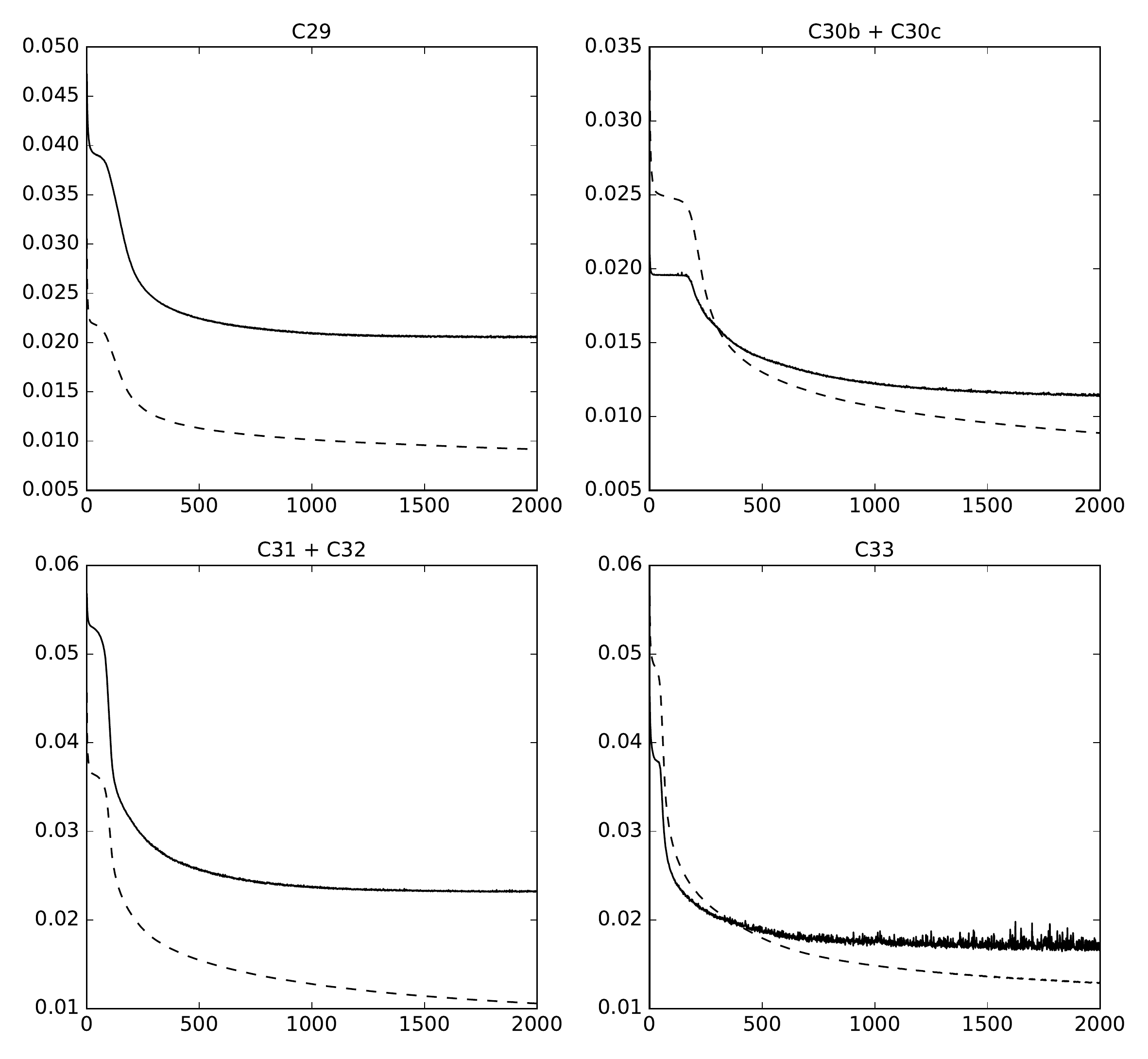}
	\caption{Training loss (dash) and validation loss (solid line)}
	\label{fig:loss}
\end{figure}

While training on each dataset, we observed that the validation loss displayed an overall tendency to decrease with the training loss, as shown in Figure \ref{fig:loss}. We have not observed any occurrence of overfitting, as the validation loss kept decreasing asymptotically. After 2000 epochs, the improvement in validation loss was negligible (of the order of $10^{-5}$), so we stopped training at that point. At about 30 seconds per epoch, the training on each dataset took about 17 hours to complete.

\begin{figure*}
	\centering
	\includegraphics[scale=0.37]{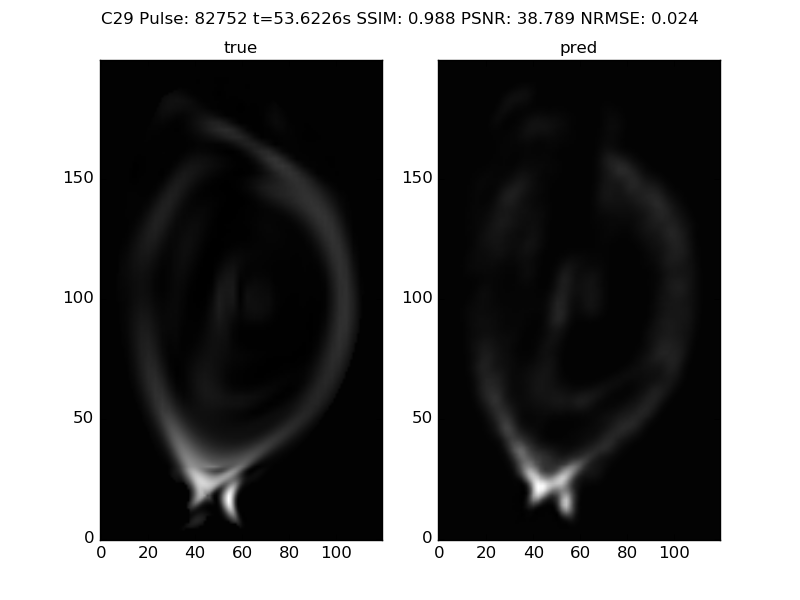}
	\includegraphics[scale=0.37]{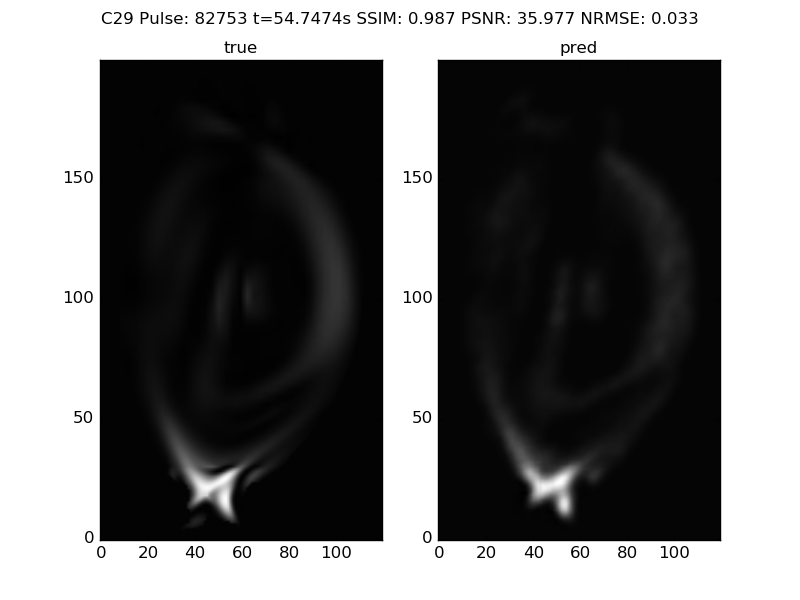}
	\includegraphics[scale=0.37]{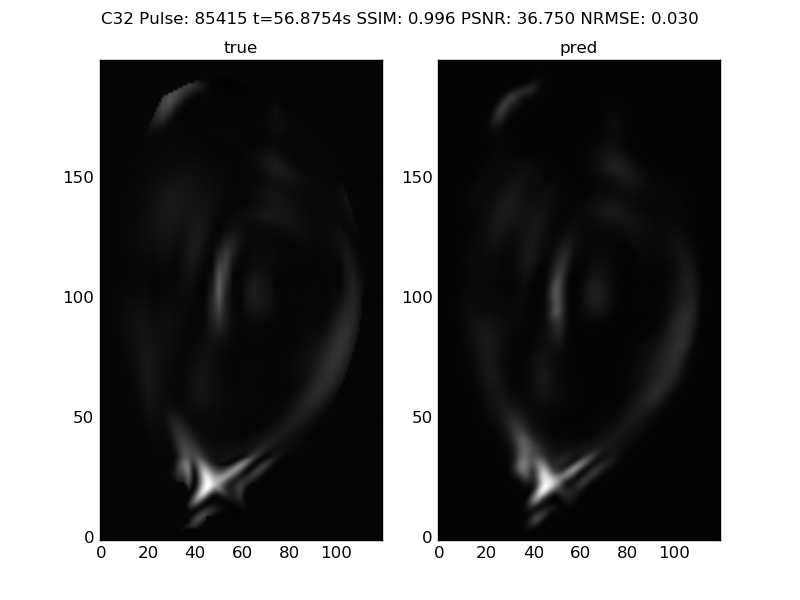}
	\includegraphics[scale=0.37]{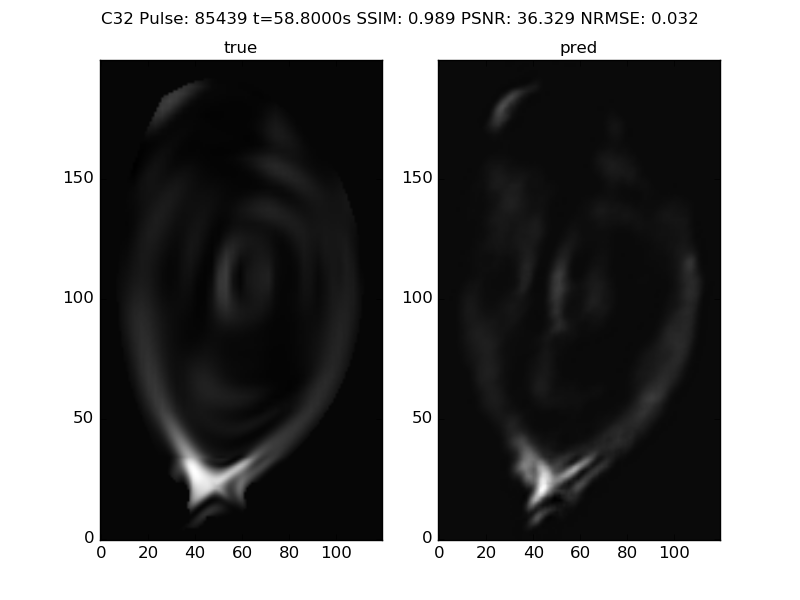}
	\includegraphics[scale=0.37]{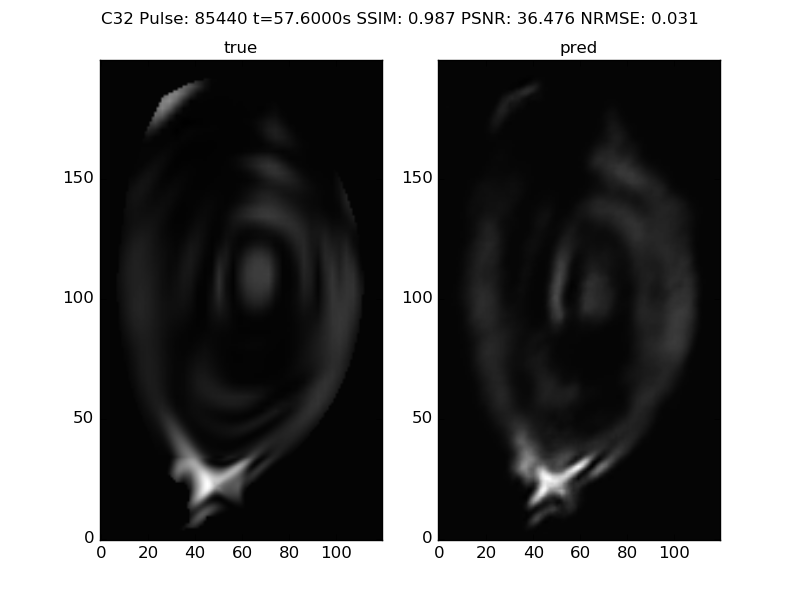}
	\includegraphics[scale=0.37]{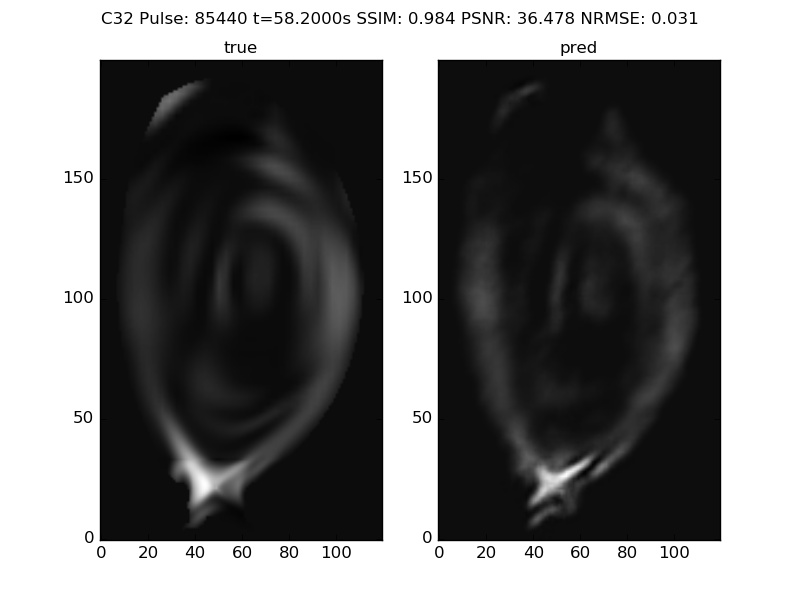}
	\includegraphics[scale=0.37]{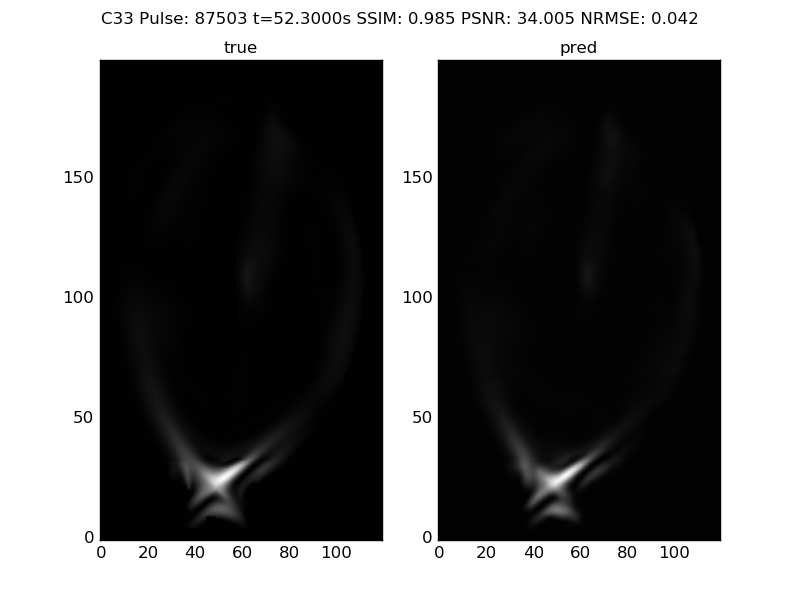}
	\includegraphics[scale=0.37]{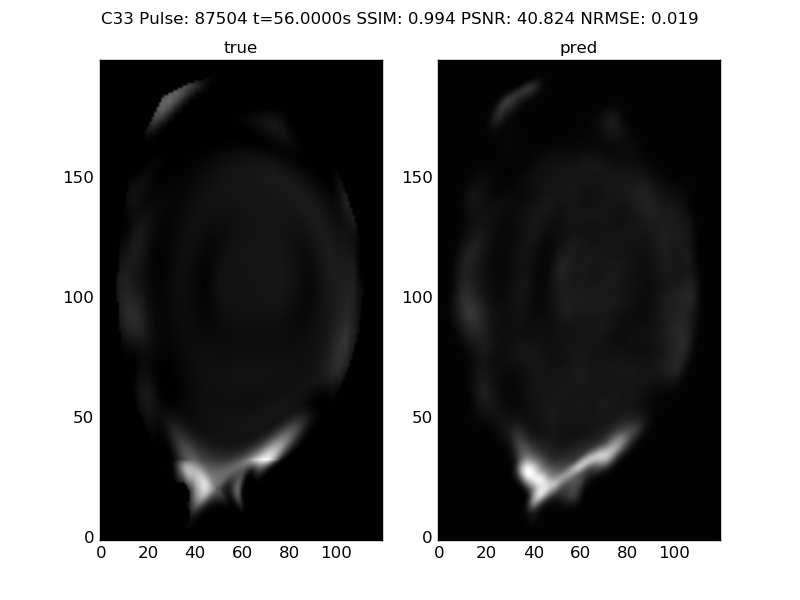}
	\caption{Sample reconstructions produced by the network, with original tomogram (left) and network output (right)}
	\label{fig:reconstructions}
\end{figure*}

We then turned to the test sets to evaluate the results on previously unseen data. Figure \ref{fig:reconstructions} shows some of the sample reconstructions obtained on the test sets. In each pair of images, the original tomogram is shown on the left, and the reconstruction produced by the network is shown on the right. The examples in Figure \ref{fig:reconstructions} indicate that the network is able to reproduce the original tomograms with high accuracy, not only with regard to the inner shape of the plasma, but also with respect to the details of the divertor region.

To make a more objective and systematic assessment of the results, we turned to several metrics that are commonly used for image comparison, namely: structural similarity (SSIM) \cite{wang04image}, which measures the perceived quality of an image with respect to another reference image; peak signal-to-noise ratio (PSNR) \cite{thu08psnr}, which is frequently used to measure the quality of image compression; and normalized root mean square error (NRMSE), where the normalization factor was the Euclidean norm of the original image. 

Table \ref{tab:results} shows the results that have been obtained for each of these metrics on each test set. From these results, we highlight the following:
\begin{itemize}

\item SSIM has a maximum value of 1.0 (only reached when the two images are equal). In two of the test sets, the average value is 0.99, which points to the high quality of the reconstructions produced by the network.

\item PSNR is logarithmic and is measured in decibels (dB). For comparison, it is worth noting that a JPEG image has typically a PSNR of 30 to 50 dB when compared to its original, uncompressed version \cite{ebrahimi04jpeg}. Here, we have values of PSNR that are in that range as well.

\item NRMSE is a measure of error and therefore the lower it is, the better. In the test sets, its average value is in the range of 0.02 to 0.04. Intuitively, one could say that, on average, the reconstruction error for each pixel is around 2\% to 4\%.

\end{itemize}

At this point, it should be noted that all tomographic reconstructions can be affected to some extent by artifacts \cite{mlynar10inversion} which are due to noise in the projection data or to the smoothing constraints that are applied during the reconstruction process. These artifacts will be present in the training data. A network that is able to reproduce existing tomograms with a very low error, as indicated by the metrics above, will necessarily learn to reproduce those artifacts as well. This problem, however, is beyond the scope of this work. Our goal here was to devise and train a network to reproduce, as faithfully as possible, the results of the reconstruction process.

\section{Conclusion}
\label{sec:conclusion}

Deep learning is a promising approach to carry out the computationally intensive task of generating a tomographic reconstruction for each reading of the KB5 diagnostic. Once trained on a set of sample tomograms, a deep neural network is capable of producing reconstructions with the same pixel resolution and with a very low error margin. This makes it possible to envision a tomography diagnostic where a trained network would be able to provide the tomographic reconstruction within the temporal constraints of real-time control systems.

The work described here is of interest not only for bolometer systems such as KB5, but also for other diagnostics based on soft X-ray cameras \cite{alper97xray} and neutron cameras \cite{adams93neutron}, which have been used for plasma tomography as well. In principle, it should be possible to adapt the up-convolutional network to different diagnostics, depending on the size of the input data and the desired resolution for the output image.

Besides the diagnostics available at JET, a similar approach could be used in other fusion devices. For example, the tomography diagnostic in Wendelstein 7-X has as many as 20 X-ray cameras with a total of 360 lines of sight \cite{schulke13technology}. A deep learning approach such as the one proposed here could bring significant savings in the time required to compute the reconstruction of the plasma profile from that amount of data.

\section*{Acknowledgements}
\small
\noindent This work has been carried out within the framework of the EUROfusion Consortium and has received funding from the Euratom research and training programme 2014-2018 under grant agreement No 633053. The views and opinions expressed herein do not necessarily reflect those of the European Commission. IPFN activities received financial support from Funda\c{c}\~{a}o para a Ci\^{e}ncia e Tecnologia (FCT) through project UID/FIS/50010/2013.


\bibliographystyle{elsarticle-num}
\bibliography{article}

\begin{thebibliography}{10}
\expandafter\ifx\csname url\endcsname\relax
  \def\url#1{\texttt{#1}}\fi
\expandafter\ifx\csname urlprefix\endcsname\relax\def\urlprefix{URL }\fi
\expandafter\ifx\csname href\endcsname\relax
  \def\href#1#2{#2} \def\path#1{#1}\fi

\bibitem{anton96xray}
M.~Anton, H.~Weisen, M.~J. Dutch, W.~von~der Linden, F.~Buhlmann, R.~Chavan,
  B.~Marletaz, P.~Marmillod, P.~Paris, X-ray tomography on the {TCV} tokamak,
  Plasma Physics and Controlled Fusion 38~(11) (1996) 1849--1878.

\bibitem{ingesson98tomography}
L.~Ingesson, B.~Alper, H.~Chen, A.~Edwards, G.~Fehmers, J.~Fuchs, R.~Giannella,
  R.~Gill, L.~Lauro-Taroni, M.~Romanelli, Soft {X} ray tomography during {ELMs}
  and impurity injection in {JET}, Nuclear Fusion 38~(11) (1998) 1675--1694.

\bibitem{ingesson08tomography}
L.~C. Ingesson, B.~Alper, B.~J. Peterson, J.-C. Vallet, Tomography diagnostics:
  Bolometry and soft-x-ray detection, Fusion Science and Technology 53~(2)
  (2008) 528--576.

\bibitem{mlynar10inversion}
J.~Mlynar, V.~Weinzettl, G.~Bonheure, A.~Murari, Inversion techniques in the
  soft-x-ray tomography of fusion plasmas: Toward real-time applications,
  Fusion Science and Technology 58~(3) (2010) 733--741.

\bibitem{kim06comparison}
J.~Kim, S.~H. Lee, W.~Choe, Comparison of the three tokamak plasma tomography
  methods for high spatial resolution and fast calculation, Review of
  Scientific Instruments 77~(10).

\bibitem{craciunescu09comparison}
T.~Craciunescu, G.~Bonheure, V.~Kiptily, A.~Murari, I.~Tiseanu, V.~Zoita, A
  comparison of four reconstruction methods for {JET} neutron and gamma
  tomography, Nuclear Instruments and Methods in Physics Research Section A
  605~(3) (2009) 374--383.

\bibitem{demeter97neural}
G.~Demeter, Tomography using neural networks, Review of Scientific Instruments
  68~(3) (1997) 1438--1443.

\bibitem{barana02neural}
O.~Barana, A.~Murari, P.~Franz, L.~C. Ingesson, G.~Manduchi, Neural networks
  for real time determination of radiated power in jet, Review of Scientific
  Instruments 73~(5) (2002) 2038--2043.

\bibitem{ronchi10neural}
E.~Ronchi, S.~Conroy, E.~A. Sundén, G.~Ericsson, M.~G. Johnson, C.~Hellesen,
  H.~Sjöstrand, M.~Weiszflog, Neural networks based neutron emissivity
  tomography at {JET} with real-time capabilities, Nuclear Instruments and
  Methods in Physics Research Section A 613~(2) (2010) 295--303.

\bibitem{carvalho10control}
P.~Carvalho, H.~Thomsen, R.~Coelho, P.~Duarte, C.~Silva, H.~Fernandes, {ISTTOK}
  plasma control with the tomography diagnostic, Fusion Engineering and Design
  85~(2) (2010) 266--271.

\bibitem{nature15deeplearning}
Y.~LeCun, Y.~Bengio, G.~Hinton, Deep learning, Nature 521~(7553) (2015)
  436--444.

\bibitem{schmidhuber15deep}
J.~Schmidhuber, Deep learning in neural networks: An overview, Neural Networks
  61 (2015) 85--117.

\bibitem{krizhevsky12imagenet}
A.~Krizhevsky, I.~Sutskever, G.~E. Hinton, {ImageNet} classification with deep
  convolutional neural networks, in: Advances in Neural Information Processing
  Systems 25, 2012, pp. 1097--1105.

\bibitem{simonyan15verydeep}
K.~Simonyan, A.~Zisserman, Very deep convolutional networks for large-scale
  image recognition, in: International Conference on Learning Representations,
  2015.

\bibitem{graves12recurrent}
A.~Graves, Supervised Sequence Labelling with Recurrent Neural Networks, Vol.
  385 of Studies in Computational Intelligence, Springer, 2012.

\bibitem{nair10rectified}
V.~Nair, G.~E. Hinton, Rectified linear units improve restricted {Boltzmann}
  machines, in: Proceedings of the 27th International Conference on Machine
  Learning (ICML-10), 2010, pp. 807--814.

\bibitem{maas13rectifier}
A.~L. Maas, A.~Y. Hannun, A.~Y. Ng, Rectifier nonlinearities improve neural
  network acoustic models, in: Proceedings of the 30th International Conference
  on Machine Learning, 2013.

\bibitem{he15delving}
K.~He, X.~Zhang, S.~Ren, J.~Sun, Delving deep into rectifiers: Surpassing
  human-level performance on {ImageNet} classification, in: Proceedings of the
  IEEE International Conference on Computer Vision, 2015, pp. 1026--1034.

\bibitem{clevert16elu}
D.-A. Clevert, T.~Unterthiner, S.~Hochreiter, Fast and accurate deep network
  learning by exponential linear units ({ELUs}), in: International Conference
  on Learning Representations, 2016.

\bibitem{glorot10understanding}
X.~Glorot, Y.~Bengio, Understanding the difficulty of training deep feedforward
  neural networks, in: 13th International Conference on Artificial Intelligence
  and Statistics, 2010, pp. 249--256.

\bibitem{duchi11adagrad}
J.~Duchi, E.~Hazan, Y.~Singer, Adaptive subgradient methods for online learning
  and stochastic optimization, Journal of Machine Learning Research 12 (2011)
  2121--2159.

\bibitem{zeiler12adadelta}
M.~D. Zeiler, Adadelta: An adaptive learning rate method, Tech. rep., Google
  (2012).

\bibitem{kingma15adam}
D.~Kingma, J.~Ba, Adam: A method for stochastic optimization, in: International
  Conference on Learning Representations, 2015.

\bibitem{srivastava14dropout}
N.~Srivastava, G.~Hinton, A.~Krizhevsky, I.~Sutskever, R.~Salakhutdinov,
  Dropout: A simple way to prevent neural networks from overfitting, Journal of
  Machine Learning Research 15~(1) (2014) 1929--1958.

\bibitem{ioffe15batch}
S.~Ioffe, C.~Szegedy, Batch normalization: Accelerating deep network training
  by reducing internal covariate shift, in: Proceedings of The 32nd
  International Conference on Machine Learning, 2015, pp. 448--456.

\bibitem{raina09gpus}
R.~Raina, A.~Madhavan, A.~Y. Ng, Large-scale deep unsupervised learning using
  graphics processors, in: Proceedings of the 26th Annual International
  Conference on Machine Learning, 2009, pp. 873--880.

\bibitem{coates13deep}
A.~Coates, B.~Huval, T.~Wang, D.~Wu, B.~Catanzaro, N.~Andrew, Deep learning
  with {COTS} {HPC} systems, in: Proceedings of the 30th international
  conference on machine learning, 2013, pp. 1337--1345.

\bibitem{jia14caffe}
Y.~Jia, E.~Shelhamer, J.~Donahue, S.~Karayev, J.~Long, R.~Girshick,
  S.~Guadarrama, T.~Darrell, Caffe: Convolutional architecture for fast feature
  embedding, in: Proceedings of the 22nd ACM International Conference on
  Multimedia, 2014, pp. 675--678.

\bibitem{bergstra11theano}
J.~Bergstra, F.~Bastien, O.~Breuleux, P.~Lamblin, R.~Pascanu, O.~Delalleau,
  G.~Desjardins, D.~Warde-Farley, I.~Goodfellow, A.~Bergeron, Y.~Bengio,
  Theano: Deep learning on {GPUs} with {Python}, in: NIPS 2011, Big Learning
  Workshop, 2011.

\bibitem{collobert11torch}
R.~Collobert, K.~Kavukcuoglu, C.~Farabet, Torch7: A {Matlab}-like environment
  for machine learning, in: NIPS 2011, Big Learning Workshop, 2011.

\bibitem{ronchi10parametric}
E.~Ronchi, S.~Conroy, E.~A. Sund\'{e}n, G.~Ericsson, M.~G. Johnson,
  C.~Hellesen, H.~Sj\"{o}strand, M.~Weiszflog, A parametric model for fusion
  neutron emissivity tomography for the {KN3} neutron camera at {JET}, Nuclear
  Fusion 50~(3) (2010) 035008.

\bibitem{huber07upgraded}
A.~Huber, K.~McCormick, P.~Andrew, P.~Beaumont, S.~Dalley, J.~Fink, J.~Fuchs,
  K.~Fullard, W.~Fundamenski, L.~Ingesson, F.~Mast, S.~Jachmich, G.~Matthews,
  P.~Mertens, V.~Philipps, R.~Pitts, S.~Sanders, W.~Zeidner, Upgraded bolometer
  system on {JET} for improved radiation measurements, Fusion Engineering and
  Design 82~(5--14) (2007) 1327--1334.

\bibitem{mast85bolometric}
K.~F. Mast, H.~Krause, K.~Behringer, A.~Bulliard, G.~Magyar, Bolometric
  diagnostics in {JET}, Review of Scientific Instruments 56~(5) (1985)
  969--971.

\bibitem{ingesson00characterization}
L.~C. Ingesson, C.~F. Maggi, R.~Reichle, Characterization of geometrical
  detection-system properties for two-dimensional tomography, Review of
  Scientific Instruments 71~(3) (2000) 1370--1378.

\bibitem{mccormick05new}
K.~McCormick, A.~Huber, C.~Ingesson, F.~Mast, J.~Fink, W.~Zeidner, A.~Guigond,
  S.~Sanders, New bolometry cameras for the {JET} enhanced performance phase,
  Fusion Engineering and Design 74~(1--4) (2005) 679--683.

\bibitem{huber07improved}
A.~Huber, K.~McCormick, P.~Andrew, M.~de~Baar, P.~Beaumont, S.~Dalley, J.~Fink,
  J.~Fuchs, K.~Fullard, W.~Fundamenski, L.~Ingesson, G.~Kirnev, P.~Lomas,
  F.~Mast, S.~Jachmich, G.~Matthews, P.~Mertens, A.~Meigs, V.~Philipps,
  J.~Rapp, G.~Saibene, S.~Sanders, R.~Sartori, M.~Stamp, W.~Zeidner, Improved
  radiation measurements on {JET} -- first results from an upgraded bolometer
  system, Journal of Nuclear Materials 363--365 (2007) 365--370.

\bibitem{ingesson97radiation}
L.~C. Ingesson, G.~C. Fehmers, H.~Y. Guo, L.~Lauro-Taroni, A.~Loarte,
  R.~Reichle, R.~Simonini, Radiation distribution and neutral-particle loss in
  the {JET} {Mk I} and {Mk IIA} divertors, in: 24th EPS Conference on
  Controlled Fusion and Plasma Physics, 1997.

\bibitem{ingesson00mathematics}
L.~C. Ingesson, The mathematics of some tomography algorithms used at {JET},
  Tech. Rep. JET-R(99)08, JET Joint Undertaking (March 2000).

\bibitem{horton13wall}
L.~Horton, The {JET} {ITER}-like wall experiment: First results and lessons for
  {ITER}, Fusion Engineering and Design 88~(6--8) (2013) 434--439.

\bibitem{jolliffe02pca}
I.~T. Jolliffe, Principal Component Analysis, 2nd Edition, Springer Series in
  Statistics, Springer, 2002.

\bibitem{lee09unsupervised}
H.~Lee, P.~Pham, Y.~Largman, A.~Y. Ng, Unsupervised feature learning for audio
  classification using convolutional deep belief networks, in: Advances in
  Neural Information Processing Systems 22, 2009, pp. 1096--1104.

\bibitem{ngiam11multimodal}
J.~Ngiam, A.~Khosla, M.~Kim, J.~Nam, H.~Lee, A.~Ng, Multimodal deep learning,
  in: Proceedings of the 28th International Conference on Machine Learning
  (ICML-11), 2011, pp. 689--696.

\bibitem{zeiler11adaptive}
M.~D. Zeiler, G.~W. Taylor, R.~Fergus, Adaptive deconvolutional networks for
  mid and high level feature learning, in: International Conference on Computer
  Vision (ICCV), IEEE, 2011, pp. 2018--2025.

\bibitem{noh15deconvolution}
H.~Noh, S.~Hong, B.~Han, Learning deconvolution network for semantic
  segmentation, in: IEEE International Conference on Computer Vision (ICCV),
  2015, pp. 1520--1528.

\bibitem{dosovitskiy15learning}
A.~Dosovitskiy, J.~T. Springenberg, T.~Brox, Learning to generate chairs with
  convolutional neural networks, in: IEEE Conference on Computer Vision and
  Pattern Recognition (CVPR), 2015, pp. 1538--1546.

\bibitem{lecun12backprop}
Y.~A. LeCun, L.~Bottou, G.~B. Orr, K.-R. M\"{u}ller, Efficient backprop, in:
  Neural Networks: Tricks of the Trade, Vol. 7700 of LNCS, Springer, 2012, pp.
  9--48.

\bibitem{wang04image}
Z.~Wang, A.~C. Bovik, H.~R. Sheikh, E.~P. Simoncelli, Image quality assessment:
  from error visibility to structural similarity, IEEE Transactions on Image
  Processing 13~(4) (2004) 600--612.

\bibitem{thu08psnr}
Q.~Huynh-Thu, M.~Ghanbari, Scope of validity of {PSNR} in image/video quality
  assessment, Electronics Letters 44~(13) (2008) 800--801.

\bibitem{ebrahimi04jpeg}
F.~Ebrahimi, M.~Chamik, S.~Winkler, {JPEG} vs. {JPEG} 2000: an objective
  comparison of image encoding quality, in: Applications of Digital Image
  Processing XXVII, Vol. 5558 of Proceedings of SPIE, 2004, pp. 300--308.

\bibitem{alper97xray}
B.~Alper, S.~Dillon, A.~W. Edwards, R.~D. Gill, R.~Robins, D.~J. Wilson, The
  {JET} soft x-ray diagnostic systems, Review of Scientific Instruments 68~(1)
  (1997) 778--781.

\bibitem{adams93neutron}
J.~Adams, O.~Jarvis, G.~Sadler, D.~Syme, N.~Watkins, The {JET} neutron emission
  profile monitor, Nuclear Instruments and Methods in Physics Research Section
  A 329~(1--2) (1993) 277--290.

\bibitem{schulke13technology}
M.~Sch\"{u}lke, A.~Cardella, D.~Hathiramani, S.~Mettchen, H.~Thomsen,
  S.~Wei{\ss}flog, D.~Zacharias, Technology development of the soft {X-ray}
  tomography system in {Wendelstein 7-X} stellarator, Fusion Engineering and
  Design 88~(9--10) (2013) 1987--1991.

\end{thebibliography}

\end{document}